# When Do Composition Errors Change Decisions? Order Effects in Frozen Representations and World Models

Thomas Serval — Baracoda AI Labs; NEOMA Business School. Preprint, November 2026.

## Abstract

Does an error in composing two learned actions change what an agent chooses? Planning with action-conditioned predictors in JEPA-style world models assumes that actions learned one at a time predict what happens when two of them are applied in either order. We measure how well this holds, and what it costs for decisions, in four frozen vision encoders (DINOv2, CLIP, SigLIP, MAE) and in two pretrained world models.

On frozen encoders, composing affine maps fitted one action at a time on initial states predicts the order gap worse than a direct read-out of the initial representation in all 24 paired contrasts (fidelity margins 0.05 to 1.26). Training each map on initial and transformed states recovers much of this deficit and cuts the regret of a two-action choice by 0.40–0.57 for scale and translation; a shared non-linear operator adds fidelity, while the identity of the previous action adds almost nothing in the sequences studied. In Push-T, two pretrained world models from the JEPA-WMs release recover about half of the latent order gap. Their planners recognise which order produced a goal image even when the block ends in the same pose, yet, with goals and costs defined on the block pose alone, among the 221 of 300 fresh scenes meeting the pose-effect threshold, observing the true intermediate state reduced block-pose regret by about a quarter for both models (JEPA-WM 0.081 [0.042, 0.125], DINO-WM 0.065 [0.022, 0.114]).

When the better order depends on an object's visible heading, choices track how linearly the heading can be read from each layer. Appendices report causal tests of weekday-shift operators in two language models and a test of operator transfer between encoders. Recognising a trajectory is not the same as predicting the consequence that matters for the goal.

# 1 Introduction

The world has an algebra. Moving an object left then up lands where moving it up then left does. Rotating it about a point and then shifting it does not. A representation that captures how the world changes should let a learner predict these order effects from what it has learned about each action alone. Planning with a learned world model rests on this: actions are learned one at a time and chained at decision time.

The idea that a good representation is one on which transformations act as a group is old (Higgins et al., 2018). It has mostly served as a training objective on toy worlds, and equivariance has been measured one transformation at a time (Lenc & Vedaldi, 2015; Gruver et al., 2023). We ask a second-order question of the frozen models in use: do separately learned actions predict when, and by how much, order matters?

The answer we find is specific.

- The order effect is linearly present in the initial representation. Where the pose is visible, a direct affine read-out predicts it well (fidelity 0.74–0.90 on fully visible photographs), well above an image-free constant (0.35–0.65).
- Composing actions learned on initial states predicts it much worse. For scale and translation on fully visible photographs it does worse than that constant (Figure 1).
- Two simple changes to how the actions are learned recover most of the deficit, for different reasons. Training each action on a mixture of initial and already transformed states fixes a coverage problem. Fitting each step on the states it actually receives adds a further gain for scale and translation.
- In a two-action choice task where the better order depends only on the goal, the first change is the one that reduces wrong choices.
- When the better order depends on which way the object faces, the bottleneck moves: the tested affine predictors choose about as well as the heading information they can read allows, and three of the four encoders keep the heading only weakly in linear form (Figure 3).

Our operators are the simplest action-conditioned predictors one can attach to a frozen encoder. Two of our encoders, DINOv2 and MAE, are the off-the-shelf encoders on which Balestriero, LeCun and Varaku (2026) build their counterfactual world models. In JEPA world models, a predictor trained on observed transitions is chained at planning time (Terver et al., 2025), and multi-step rollout losses, as in V-JEPA 2-AC (Assran et al., 2025), feed it its own predictions during training. Our experiments test a related but different thing: training on real intermediate states, encoded from the transformed images, improves some affine compositions. Predicted and encoded intermediate states follow different distributions, so whether rollout training brings the same benefit is left open.

The trigger for this work is Balestriero, LeCun and Varaku's Counterfactual World Models (2026), whose counterfactual rule fails at a second-order coupling between an action and a disturbance. Section 10 shows that this coupling has a symmetric part, which their analysis bounds, and an antisymmetric part, the commutator, which the order gap measures. Section 9 extends the convergence question of Balestriero and LeCun's Platonic Representation Theorem from points to operators.

Contributions.

- A protocol for frozen representations: four normalised algebraic defects, four predictors of the order gap (direct, composed, shared, specialised), validation-selected layers, paired bootstrap intervals, and exact positive and negative controls (Sections 3 and 4).
- The main result. In 24 of 24 paired contrasts, composing initial-state operators predicts the order gap worse than a direct read-out. The deficit splits into coverage and conditioning on the preceding action, in proportions that depend on the transformation pair and the setting (Section 5).
- A two-action choice test that links the representation-level diagnostic to choices scored in the world (Section 6).
- Layer profiles separating decodability from compositional fidelity (Section 7), causal tests in two language models (Section 8), a test of transfer between encoders, set against the Platonic Representation Theorem (Section 9), and a pilot on two pretrained Push-T world models, where trajectory recognition is near perfect but the intermediate error still costs a quarter to a third of the regret on block goals (end of Section 6).

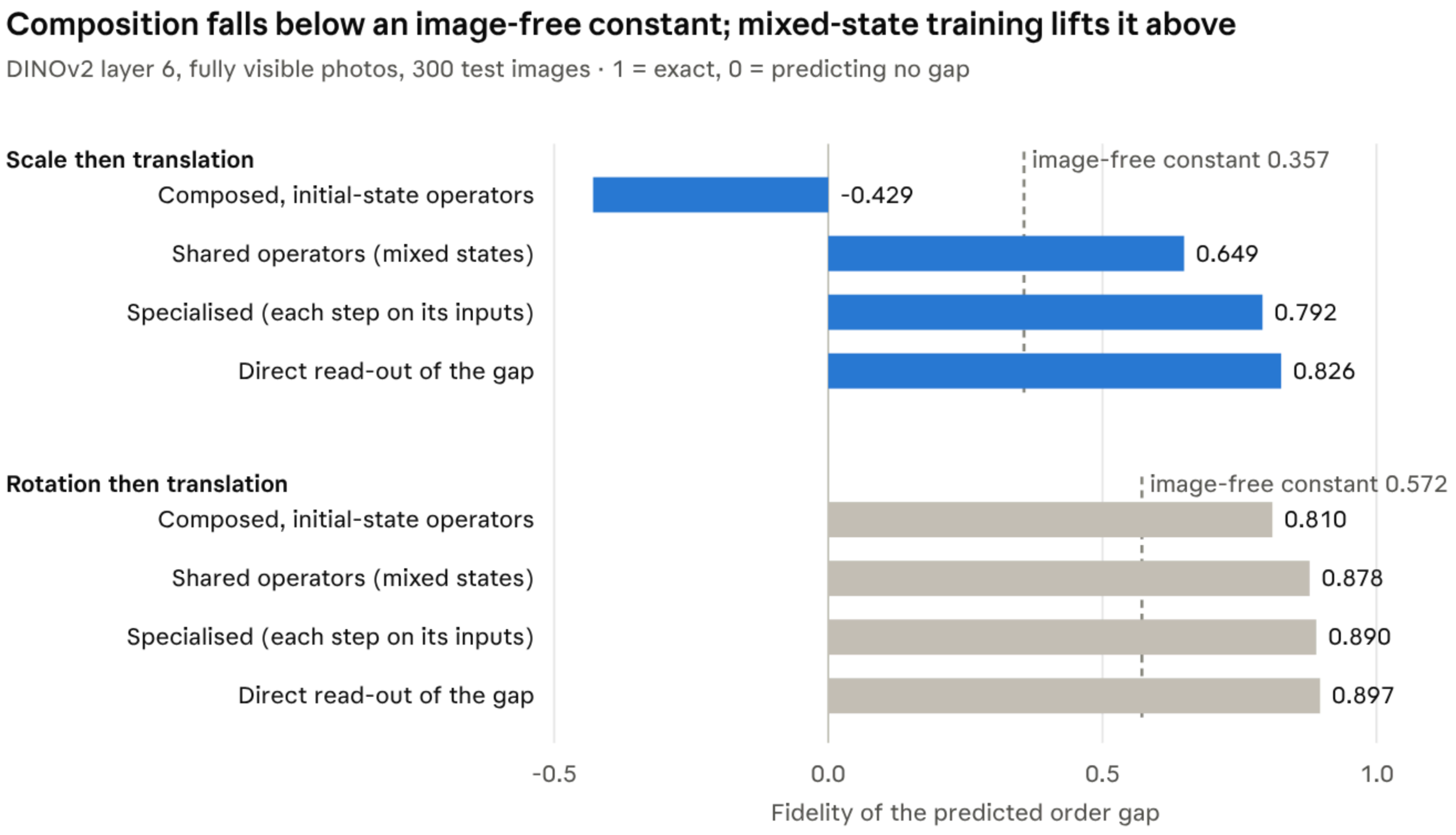

Figure 1 · results/review2_disc2000.json, DINOv2 best layer on validation, 1,700 training and 300 test photographs

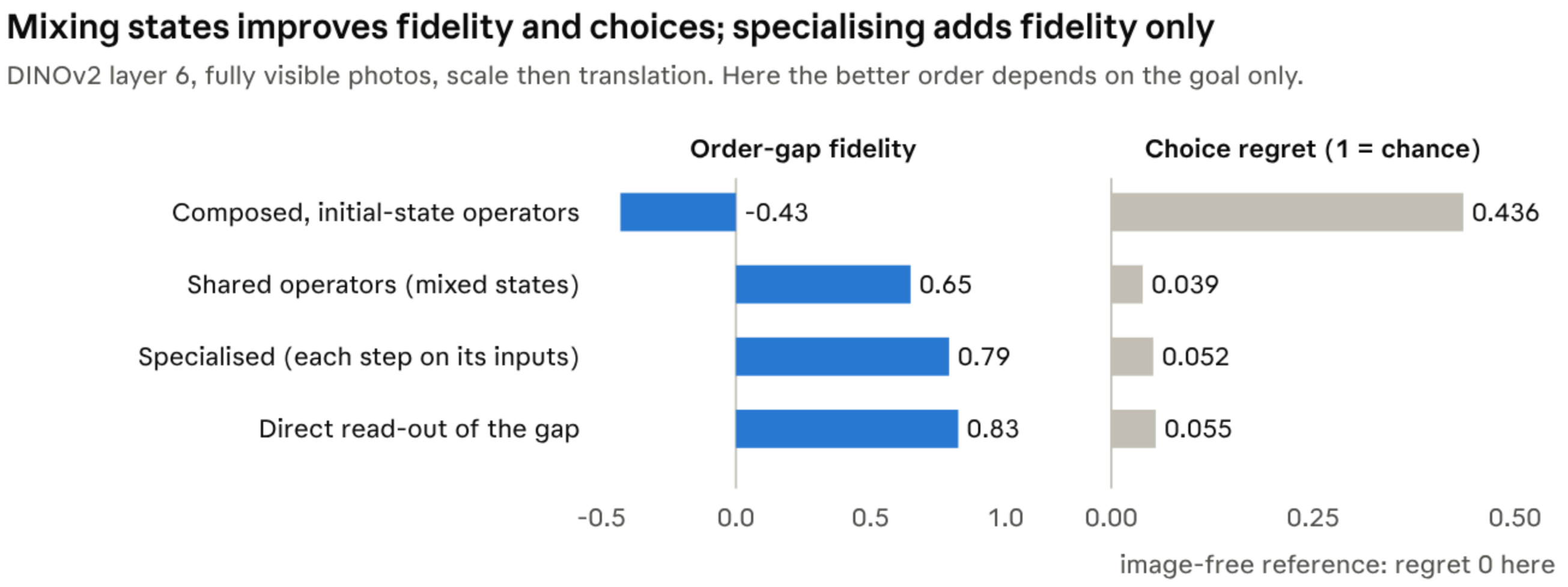

Figure 1b · results/review2_disc2000.json and decision_disc2000.json, DINOv2 L6, 300 test photographs × 16 goals

# 2 Related work

| Line of work | What it does | What it leaves open |
|---|---|---|
| Symmetry-based and homomorphism representations (Higgins et al., 2018; Keurti et al., 2023; Dean et al., 2025) | Impose or learn a group action in latent space | No measurement on frozen foundation models |
| Equivariance of learned representations (Lenc & Vedaldi, 2015; Gruver et al., 2023; Chen et al., 2026) | Learn maps between transformed representations; first-order equivariance | No composition, no order effect |
| Group-law tests on world models (Wang et al., 2026) | Identity, inverse and composition in video world models | No commutator, no language models, no comparison across encoders |
| Geometry of language models (Park et al., 2024; Engels et al., 2025; Todd et al., 2024) | Linear, circular and additive structure; circular weekday features with causal interventions (Engels et al.) | Sums, not order; no test of separately learned shifts against a known algebra |
| Order effects (Doan et al., 2026; Sweeney, 2026) | Composition of steering interventions; commutators of training stages | Not a fidelity test of a representation |
| Convergence (Huh et al., 2024; Gröger et al., 2026; Balestriero & LeCun, 2026; Koepke et al., 2026) | Similarity of points and neighbourhoods across models | Similarity of transformations |

Engels et al. (2025) are the direct predecessor for the weekday world: they show circular day features and causal calendar interventions. Our contribution there is a quantitative test of separately learned shifts, of their composition and of their closure. Lenc and Vedaldi (2015) learn maps between transformed representations of a CNN; we add the order effect, decodability controls and interventions.

Measuring non-commutativity in a learned latent space is not new in itself. Dehdarirad et al. (2026) measure latent non-commutativity in a Lie-group VAE, swap the order of transformations and study its effect on reconstruction, observing a mismatch between latent non-commutativity and its consequences at the output. Our setting differs: the actions of the world are known, the encoder is frozen, both outcomes are observed on the same scenes, and we compare several ways of predicting their difference before measuring the cost of a wrong choice. Closest on decisions, Qiu et al. (2026) separate prediction accuracy from discrimination between candidate actions and planning regret; Terver et al. (2025) study what drives planning success in JEPA world models. Koopman theory asks when chosen observables let a dynamics act as a matrix (Brunton et al., 2016), the question behind our affine operators.

# 3 Framework

Setting. A world state $w$ produces an observation $x = o(w)$. A world transformation $g$ acts on states, and $T_g$ is its effect on observations: $T_g(o(w)) = o(g \cdot w)$. A frozen encoder $f$ maps an observation to features $z = f(x)$. In the visual worlds the transformations are direct similitudes of the plane, the group $\mathbb{R}^2 \rtimes (SO(2) \times \mathbb{R}_{>0})$. Two translations commute, as do a rotation and a scaling about the same point. Rotation with translation and scale with translation do not commute.

Features. Each layer's features are reduced by PCA (128 dimensions for vision, 64 for language models) fitted on training images. They are then centred and divided by one global scale computed on the untransformed training states, which keeps the representation's own Euclidean geometry. An earlier version of this work standardised each PCA coordinate separately. On PCA coordinates that is whitening: it gives noise-dominated directions the weight of principal ones. It lowered every score by 0.1 to 0.4 and broke the pixel control on photographs with a plain background (Appendix B). All results here use the global scale.

Operators. For each transformation $g$ we fit an affine operator by ridge regression on pairs $(f(x), f(T_g x))$, with the penalty chosen on a validation split:

```
\hat f(T_g x) = A_g\, f(x) + b_g, \qquad \tilde A_g = \begin{pmatrix} A_g &
b_g \\ 0 & 1 \end{pmatrix}
```

The homogeneous form makes composition a matrix product. One-step $R^2$ is the held-out $R^2$ of the predicted change $f(T_g x) - f(x)$.

Four defects. All are ratios of squared norms (energies) on held-out data. The reference value differs per defect: identity operators give 0 for the inverse defect and for the predicted commutator on commuting pairs, and 1 for composition.

| Defect | Prediction tested | Normaliser | Exact value |
|---|---|---|---|
| Inverse | $A_{g^{-1}} A_g f(x)$ vs $f(x)$ | $E\|f(T_g x) - f(x)\|^2$ | 0 |
| Composition | $A_h A_g f(x)$ vs $f(T_h T_g x)$ | $E\|f(T_h T_g x) - f(x)\|^2$ | 0 |
| Predicted commutator, pairs that commute in the world | $(A_h A_g - A_g A_h) f(x)$ | $E\|f(T_h T_g x) - f(x)\|^2$ | 0 |
| Commutator fidelity, pairs that do not commute | $(A_h A_g - A_g A_h) f(x)$ vs the observed gap $f(T_h T_g x) - f(T_g T_h x)$ | $E\|\text{observed gap}\|^2$ | 1 (reported as 1 minus the ratio) |

The observed order gap is exactly zero for any deterministic encoder when the world commutes, since both orders give the same image. Commutation in the world is therefore not a property of the model, but whether the separately fitted operators predict the right gap is. A defect whose normaliser is zero is reported as undefined.

Four predictors of the order gap. For a non-commuting pair $(g, h)$, the target is the observed gap $z_{hg} - z_{gh}$. All four predictors are affine ridge maps fitted on the same training scenes:

- direct: one map from $z_0$ to the gap;
- composed: $A_h A_g - A_g A_h$, with both operators fitted on initial states (the commutator-fidelity defect);
- shared: the same composition, with one operator per action fitted on a balanced mixture of initial and transformed states, with the same number of transitions;
- specialised: the second step fitted on states already transformed by the first action, $A_{h|g} A_g - A_{g|h} A_h$, which conditions the second map on the preceding action.

Supervision differs between predictors, and it shapes the comparison.

- **Direct** sees, for each training scene, both final outcomes of the pair, so it receives explicit supervision of the order effect. It is a reference for how predictable the effect is from $z_0$, not a competing world model.
- **Composed** sees only elementary transitions out of initial states.
- **Shared** and **specialised** also see transitions out of the intermediate states of the two-step paths.

Per pair, with n training scenes (1,700 photographs or 350 texture scenes), direct fits one affine map on n scene pairs. Composed fits two maps on n transitions each. Shared fits two maps on n transitions each, half from initial and half from intermediate states. Specialised fits four maps: the two initial-state operators for the first step, and two second-step operators on n intermediate states each. In 128 dimensions each affine map has 129 × 128 = 16,512 parameters.

Shared minus composed measures what coverage of intermediate states recovers. Specialised and shared differ in both steps: specialised keeps the initial-state operator for the first step, whose inputs are initial states, and fits the second step on intermediate states. To isolate the second step we also report a hybrid, which uses the shared operator for the first step and the specialised operator for the second. Neither comparison shows that the representation stores the history: the preceding action may simply identify a region of latent space where a local affine map works better. Telling the two apart would require trajectories that reach the same observable state by different routes.

Image-free reference. The mean order gap over the training scenes, used as a constant prediction for every test scene, shows how much of each score needs no information about the test image.

Statistics. Layers are selected on validation. Operators are fitted on the first 80% of training scenes, and layers are ranked by composed fidelity on rotation and translation on the last 20%. All test-set comparisons are paired, with 95% bootstrap intervals over test scenes (1,000 resamples for vision, 2,000 for language models). Ranges quoted across encoders are ranges, never intervals.

What the comparison can and cannot show. For a fixed action pair, the difference between the two composed affine predictions is itself an affine function of the initial representation. A direct least-squares predictor therefore provides a target-specific reference: in the population, with the same quadratic loss and no penalty, it can only do as well or better. The sign of the held-out contrast is thus not the finding. The empirical questions are how large the held-out composition deficit is, how much of it can be recovered by changing the training states, and whether that recovery improves action choice. The comparison between direct multi-step prediction and iterating one-step models is classical in forecasting (Marcellino et al., 2006) and model-based reinforcement learning (Asadi et al., 2019); here it is applied to the difference between two orders.

# 4 Validation of the measurement

Synthetic world. The images are 64×64 renderings of an asymmetric “F” with position, orientation and scale, on a plain or a static textured background. Translations are 4 px; rotation (0.6 rad) and scaling (1.25) act about the image centre. There are 4,000 training and 1,000 test states, each rendered under every transformation and pair order. The positive control is the exact homogeneous state (x, y, cos θ, sin θ, log scale) under a random linear embedding in 256 dimensions. The negative control shuffles the training pairs. We also probe pixels (PCA 256), a random MLP, and a small CNN trained from scratch to regress the state (final training MSE 0.0025 on standardised targets).

| Representation (plain background) | One-step $R^2$ | Inverse (median) | Composition | Predicted commutator, commuting pairs | Fidelity r,tx | Fidelity sc,tx |
|---|---|---|---|---|---|---|
| Exact state, linear embedding (noise 0 and 0.05) | 1.00 | 0.000 | 0.000 | 0.000 | 1.00 | 1.00 |
| Shuffled pairs | −6.36 | 2.56 | 1.70 | 0.66 | −1.36 | −22.3 |
| Pixels PCA-256 | 0.98 | 0.035 | 0.022 | 0.010 | 0.96 | 0.79 |
| Random MLP | 0.96 | 0.121 | 0.091 | 0.146 | 0.80 | 0.06 |
| CNN layer 1 | 0.97 | 0.046 | 0.027 | 0.016 | 0.93 | 0.74 |
| CNN layer 2 | 0.96 | 0.036 | 0.034 | 0.020 | 0.85 | 0.50 |
| CNN layer 3 | 0.95 | 0.060 | 0.042 | 0.033 | 0.77 | 0.17 |
| CNN layer 4 (dense) | 0.94 | 0.095 | 0.071 | 0.069 | 0.59 | −0.85 |

Composition is the mean over the four tested pairs. The measurement passes its gates: the exact representation scores perfectly, robustly to noise, and shuffled pairs fail strongly. A network that reads out the state almost perfectly loses compositional fidelity with depth. On scale and translation its fidelity falls from 0.74 to −0.85 while one-step $R^2$ stays at 0.94–0.97. The pattern repeats on the textured background (CNN fidelity r,tx 0.90 to 0.54, sc,tx 0.69 to −0.79; pixels 0.91 and 0.70) and with a larger scale step of 1.5 (CNN sc,tx 0.76 to −1.12; pixels 0.82). Decodability does not certify compositional structure (Barenholtz, 2026).

An exactly affine control at the real sample sizes. A gap between direct and composed prediction could arise from finite samples, noise or ridge shrinkage, even in an exactly affine representation. We built one with each encoder's own spectrum:

- the exact pose state, scaled to the encoder's top five PCA variances;
- transformation-invariant scene content with the encoder's remaining spectrum;
- image-level noise from 0 to 1 times the mean variance;
- the real sample sizes (1,700 training and 300 test photographs; 350 and 150 texture scenes).

Direct and composed prediction then agree within 0.01–0.02 on photographs at one-step $R^2$ 0.86–0.92. At every noise level, composition is equal to or slightly better than direct prediction, so the finite-sample bias runs against our finding.

Ridge. With the penalty fixed from $10^{-6}$ to 1 for every map, through the near least-squares end, the real direct − composed gaps are unchanged to three decimals (for example DINOv2 on textures: 0.146–0.147 for r,tx and 0.340–0.341 for sc,tx). At $10^2$ they widen slightly.

## 5 Main result: composition versus direct prediction

Worlds.

- **Colour textures.** The F-shaped object carries its own colour texture in object coordinates and moves on static 1/f colour noise; 224×224 images, 350 training and 150 test scenes.
- **Fully visible photographs.** Each of 2,000 Imagenette photographs (Howard, 2019) is rendered as a disc of radius 36 px at unit scale on a plain background, with 30 px translations. The region needed by both paths stays in view (on average 0.1% of the disc leaves the image), and the pose is identifiable.
- **Window photographs.** A 224×224 camera window over the photograph, with 50 px translations. Content enters and leaves the frame and the absolute pose is not identifiable: a partial-observation condition.

In both photograph settings, 1,700 photographs are used for fitting and 300 disjoint ones for test. The encoders are DINOv2-base (Oquab et al., 2024), CLIP ViT-B/16 (Radford et al., 2021), SigLIP ViT-B/16 (Zhai et al., 2023) and MAE ViT-B/16 (He et al., 2022; masking off). Each is read at its spatial patch-token map, pooled to 8×8, at every second layer.

Figure 1 shows the clearest case: DINOv2 at layer 6 on fully visible photographs. Its scale-translation order gap is predicted with fidelity 0.826 directly, 0.792 with each step fitted on its own inputs, 0.649 with shared operators, 0.357 by the image-free mean training gap, and −0.429 by composing initial-state operators. Composing the usual initial-state operators therefore does worse than a constant, and training on mixed states takes it well above that constant. The same encoder, at the same layer, contains a strongly predictable order effect that the usual way of learning actions recovers very poorly. Figure 2 and the table below show that this case is representative in sign, and that all 24 contrasts exclude zero.

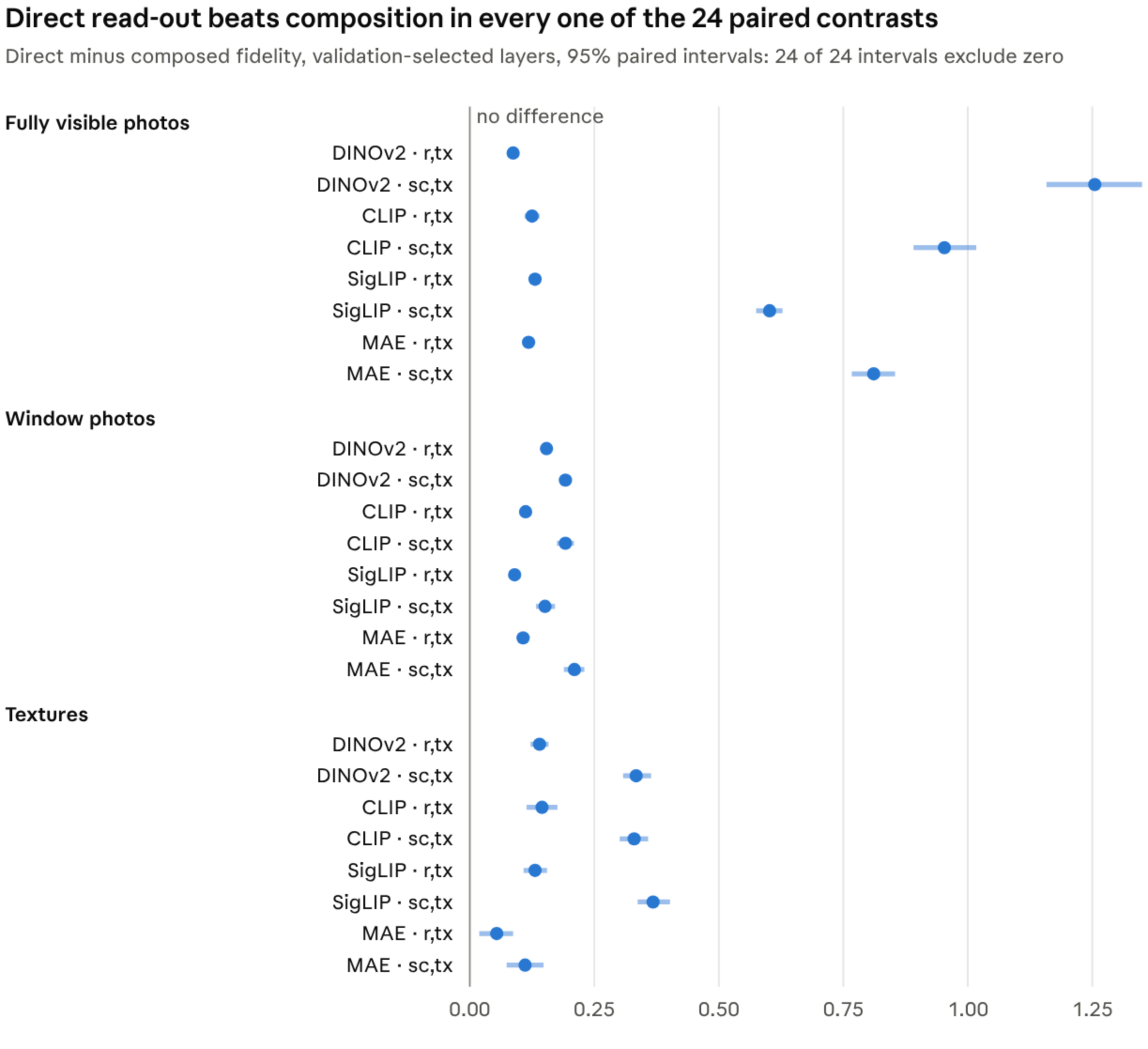

Figure 2 · results/review2_{disc2000,window2000,texture}.json, 4 encoders × 3 settings × 2 pairs

Each cell below lists five fidelities in the order: image-free mean / direct / composed / shared / specialised.

| Setting | Encoder (layer) | r,tx | sc,tx |
|---|---|---|---|
| Fully visible photos | DINOv2 (L6) | 0.572 / 0.897 / 0.810 / 0.878 / 0.890 | 0.357 / 0.826 / −0.429 / 0.649 / 0.792 |
| Fully visible photos | CLIP (L6) | 0.651 / 0.904 / 0.779 / 0.882 / 0.897 | 0.443 / 0.837 / −0.116 / 0.679 / 0.808 |
| Fully visible photos | SigLIP (L4) | 0.519 / 0.851 / 0.719 / 0.797 / 0.834 | 0.346 / 0.736 / 0.134 / 0.553 / 0.701 |
| Fully visible photos | MAE (L8) | 0.548 / 0.777 / 0.659 / 0.750 / 0.768 | 0.406 / 0.783 / −0.028 / 0.592 / 0.752 |
| Window photos | DINOv2 (L2) | 0.000 / 0.431 / 0.276 / 0.284 / 0.327 | 0.000 / 0.296 / 0.104 / 0.088 / 0.154 |
| Window photos | CLIP (L0) | 0.000 / 0.561 / 0.448 / 0.453 / 0.477 | 0.000 / 0.413 / 0.221 / 0.217 / 0.263 |
| Window photos | SigLIP (L0) | 0.000 / 0.510 / 0.420 / 0.425 / 0.445 | 0.000 / 0.381 / 0.230 / 0.222 / 0.266 |
| Window photos | MAE (L0) | 0.000 / 0.638 / 0.531 / 0.534 / 0.553 | 0.000 / 0.529 / 0.319 / 0.311 / 0.361 |
| Textures | DINOv2 (L6) | 0.435 / 0.707 / 0.567 / 0.601 / 0.706 | 0.197 / 0.544 / 0.210 / 0.245 / 0.535 |
| Textures | CLIP (L4) | 0.445 / 0.600 / 0.456 / 0.493 / 0.591 | 0.157 / 0.479 / 0.148 / 0.171 / 0.464 |
| Textures | SigLIP (L4) | 0.487 / 0.724 / 0.593 / 0.634 / 0.718 | 0.222 / 0.614 / 0.245 / 0.300 / 0.598 |
| Textures | MAE (L2) | 0.158 / 0.157 / 0.103 / 0.125 / 0.053 | 0.039 / 0.075 / −0.035 / −0.028 / −0.053 |

Against the image-free constant. Direct prediction beats the constant in 23 of the 24 contrasts, with paired intervals excluding zero (lower bounds from +0.004 to +0.619). The exception is MAE on textures for rotation and translation. There direct prediction (0.157) does no better than the constant (0.158; interval of the difference [−0.026, +0.025]), so what it captures is the average order effect, not a scene-specific one. In the window design the constant predicts nothing (fidelity 0.000 in all eight contrasts), while direct prediction reaches 0.30–0.64. Even when the absolute pose cannot be identified, the initial image carries usable information about how the two orders will change its representation. For scale and translation on fully visible photographs, composed prediction falls below the constant in all four encoders (intervals of composed minus constant between −0.875 and −0.190), and shared operators lift it above (shared minus constant +0.170 to +0.310).

Three situations, not two. Comparing the operator families separates three regimes.

- **Fully visible photographs, rotation and translation: coverage.** Shared operators recover most of the gain. Specialised exceeds shared by only 0.012 to 0.036 depending on the encoder (intervals within [+0.011, +0.039]). With the first step held at the shared operator, the second-step gain is +0.008 to +0.025 (intervals within [+0.008, +0.027]).
- **Fully visible photographs, scale and translation: coverage and step-specific fitting together.** Shared operators recover a large part of the deficit (shared − composed +0.419 to +1.078 across encoders). Fitting each step on its own inputs adds a further 0.129 to 0.159 in all four encoders (intervals within [+0.118, +0.170]). About half of that comes from the second step alone: with the first step held at the shared operator, the specialised second step adds 0.054 to 0.076 (intervals within [+0.046, +0.086]). The rest comes from keeping the initial-state operator for the first step, whose inputs are initial states.
- **Window photographs: neither helps much.** Shared operators barely improve on composition (shared − composed intervals between −0.023 and +0.013), and specialisation adds 0.018 to 0.066. Direct prediction stays well above all three. For CLIP on rotation and translation, composition gives 0.448, shared 0.453 and specialised 0.477, against 0.561 for direct prediction. When content enters and leaves the frame, the order information is in the initial state but not in a form that any of the three operator families exploits.

On textures, DINOv2, CLIP and SigLIP behave like the scale case above, with a larger gain from fitting each step on its own inputs. Specialised exceeds shared by +0.084 to +0.105 for rotation and translation, and by +0.290 to +0.298 for scale and translation; the second step alone contributes +0.085 to +0.122 of the latter (hybrid minus shared, intervals within [+0.045, +0.149]). MAE is an instructive exception. Its operators are weak at every layer on textures (one-step $R^2$ at most 0.43), and its direct prediction does no better than the image-free constant for rotation and translation. Its specialised second step does worse than the shared one (0.053 against 0.125; interval of the difference [−0.117, −0.028]): with weak operators, fitting the second step on fewer, already transformed states overfits.

History or non-linearity? The gain of step-specific fitting admits two readings: the second action needs to know what came before, or one non-linear rule acts differently in different regions of latent space. We gave a shared operator a serious chance on fully visible photographs, at the validation layers, with the same transitions and scene split for every model (initial and intermediate transitions of both actions; early stopping on held-out scenes; three seeds). A two-layer MLP f(z, action), shared by both steps, was compared with the same MLP given in addition the identity of the previous action.

| Encoder | Pair | Second-step $R^2$ from the true intermediate state: affine shared / affine specialised / MLP shared / MLP + previous action | Free-rollout fidelity: same four |
|---|---|---|---|
| DINOv2 (L6) | sc,tx | 0.923 / 0.929 / 0.932 / 0.933 | 0.660 / 0.792 / 0.766 / 0.771 |
| CLIP (L6) | sc,tx | 0.931 / 0.936 / 0.945 / 0.945 | 0.691 / 0.808 / 0.817 / 0.822 |
| SigLIP (L4) | sc,tx | 0.871 / 0.875 / 0.873 / 0.873 | 0.565 / 0.701 / 0.633 / 0.638 |
| MAE (L8) | sc,tx | 0.890 / 0.895 / 0.891 / 0.892 | 0.606 / 0.752 / 0.658 / 0.668 |

The previous action adds 0.000 to 0.014 in fidelity and at most 0.002 in second-step $R^2$ (all seeds, all eight encoder–pair combinations, rotation–translation included). The benefit of non-linearity varies: the shared MLP exceeds the specialised affine fit for CLIP (0.817 against 0.808), comes close for DINOv2 (0.766 against 0.792), and leaves a visible gap for SigLIP (0.633 against 0.701) and MAE (0.658 against 0.752). For the tested models, the identity of the previous action therefore carries little predictive information beyond the current state; in the sequences studied, the step-specific gain reflects the limits of a single affine approximation more than missing history. The scope is limited: each pair of actions is fitted separately, so at the second step the previous action is always the other action of the pair. Concluding on memory of past actions in general would need one action to follow several antecedents (A then B, C then B, B then B). Choices do not improve either: every model errs on 2.7–8.5% of scale–translation choices.

Not inherited from a poorer one-step fit. On textures and window photographs, the one-step $R^2$ of the second step is nearly the same across operator families (for example DINOv2 on textures, scale operator: 0.77 initial, 0.78 shared and specialised). On fully visible photographs, the scale operator applied to translated states improves from 0.85–0.89 (initial) to 0.91–0.96 (shared and specialised); that improvement is what the coverage gain reflects. The gap predicted by composition is mostly misdirected rather than mis-sized. For scale and translation on fully visible photographs, its median cosine with the observed gap is 0.23–0.42 while its amplitude ratio is 0.67–0.97, so it scores below a zero prediction because of its direction.

Pixels pay part of the price too. Low-resolution pixels (24×24, grey) show the same contrast:

- fully visible photographs: direct 0.95 against composed 0.83 for rotation and translation, 0.80 against 0.46 for scale and translation;
- textures: 0.45 against 0.10.

Part of the deficit is therefore a property of global affine operators in any coordinates that are not exactly equivariant. On fully visible photographs the encoders pay more for scale and translation than pixels do (composed −0.43 to 0.13, against 0.46 for pixels), while their direct prediction is comparable (0.74–0.84, against 0.80).

Output layers. At the last layer of the fully visible setting, direct prediction remains good for three encoders (r,tx 0.58–0.76, sc,tx 0.65–0.79; SigLIP 0.31 and 0.24), and composition is again worse (r,tx 0.22–0.64; sc,tx −0.76 to −0.08). On textures and window photographs, direct prediction at the output is itself weak (0.04–0.41). There, the order effect is not affinely accessible from the representation at that layer.

## 6 Consequence of the composition defect for a two-action choice

Fidelity scores the whole latent gap. A planner needs less: which of the two orders brings it closer to its goal. We therefore measured how often each predictor picks the better order.

Protocol.

- **Task.** For each held-out scene and each of 16 goal directions $u$ fixed in advance, the goal is to place the object's centre as far as possible along $u$. The agent chooses between “g then h” and “h then g”.
- **Ground truth.** The simulator gives the centre reached by both orders, recovered exactly from the noiseless state embedding (error below $4{\cdot}10^{-6}$ px).

- **Decoder.** Each predictor's latent gap is turned into a predicted displacement by one linear position decoder per representation. It is fitted on all training images and shared by all methods.
- **Metrics.** We report the error rate and the regret. The regret is the projected distance lost when the wrong order is chosen, relative to the regret of choosing at random (0 is perfect, 1 is chance). A goal perpendicular to the physical gap makes both orders equally good, and its regret is zero whatever the choice. No goal is exactly perpendicular here: the 16 directions are offset by 0.1 rad from the axes.
- **Statistics.** Intervals resample scenes with their 16 goals together.
- **Layers** are those fixed on validation in Section 5.

Controls. The witness decodes the two true final latents and compares them. The image-free reference ignores the test image. It takes the mean gap between the two final positions on the training scenes and chooses the order by its projection on the goal. Pixels cannot be tested, because the linear position decoder fails on them ($R^2$ below zero).

A property of these worlds sets what the test can show. Both actions act about the image centre, so for a given pair the physical gap of the object's centre is the same for every scene: $(I - R)t$ for rotation and translation, $(1 - s)t$ for scale and translation. The goal direction can change the better order, but the starting image never does. The image-free reference therefore makes every choice correctly from training data alone. The experiment measures how reliably each predictor recovers a task-relevant displacement that is common to varied visual inputs; it is not a planning problem. This also explains why direct prediction can err less than the witness. It can learn the common displacement, while subtracting two decoded positions keeps part of both decoding errors.

| Setting and method | r,tx: error rate | r,tx: relative regret | sc,tx: error rate | sc,tx: relative regret |
|---|---|---|---|---|
| Image-free reference (mean training gap) | 0.0% | 0.000 | 0.0% | 0.000 |
| Fully visible photos: witness, two true final latents | 2.2–3.2% | 0.013–0.020 | 6.1–9.2% | 0.068–0.094 |
| Fully visible photos: composed, initial-state operators | 10.1–14.2% | 0.082–0.127 | 24.2–29.8% | 0.436–0.581 |
| Fully visible photos: shared operators | 0.5–2.0% | 0.003–0.012 | 3.0–7.3% | 0.017–0.077 |

| Setting and method | r,tx: error rate | r,tx: relative regret | sc,tx: error rate | sc,tx: relative regret |
|---|---|---|---|---|
| Fully visible photos: specialised second step | 0.4–1.1% | 0.002–0.007 | 3.8–8.0% | 0.041–0.076 |
| Fully visible photos: direct | 0.3–1.5% | 0.002–0.009 | 3.9–8.0% | 0.049–0.077 |
| Textures: witness, two true final latents | 8.5–16.5% | 0.083–0.227 | 24.2–36.0% | 0.422–0.696 |
| Textures: composed | 3.3–11.7% | 0.020–0.119 | 2.2–49.7% | 0.007–1.011 |
| Textures: shared | 4.1–10.2% | 0.025–0.096 | 5.0–32.3% | 0.024–0.591 |
| Textures: specialised | 4.4–9.3% | 0.031–0.091 | 19.8–28.9% | 0.343–0.516 |
| Textures: direct | 4.1–5.8% | 0.027–0.038 | 19.1–25.0% | 0.325–0.439 |

Ranges are across the four encoders. On fully visible photographs the witness succeeds, so the decoder does not limit the test there.

- **The shared operator improves both metrics.** On fully visible photographs, fitting a shared action operator on both initial and transformed states reduced normalised decision regret by 0.40–0.57 for scale and translation, and by 0.08–0.12 for rotation and translation. Every paired interval excludes zero.
- **Step-specific fitting moves fidelity, not choices.** It improved latent order-gap fidelity by a further 0.13–0.16 for scale and translation, with little additional change in observed regret for the tested goals: the paired intervals of specialised minus shared regret lie between −0.015 and +0.047. For CLIP it improves fidelity and slightly increases regret relative to the shared operator (+0.024 [+0.005, +0.047]). Taking a regret reduction of 0.05 of chance level as the smallest useful gain, every interval excludes a useful gain from specialisation. Both metrics reward the first improvement; only the latent metric clearly distinguishes the second.
- **Direct prediction does not choose better than the specialised step** (differences −0.000 to +0.009).

On textures the witness errs on 8.5–16.5% of choices for rotation and translation and on 24–36% for scale and translation. With the true final states that hard to compare, the ranking of methods cannot be attributed to the operators, and we report the texture rows with that limit. For scale and translation, composed operators happen to choose best for three encoders (2–4% errors); we do not interpret this.

## When the better order depends on the image: move forward, then turn

In the test above the better order never depends on the image. We therefore built a world in which it does. Two actions act on the object itself:

- **fw** moves it along its own heading, by 14 px on textures and 30 px on fully visible photographs;
- **pv** turns it in place by 90°.

Moving then turning ends at $p + d\,u(\theta)$; turning then moving ends at $p + d\,u(\theta + 90°)$. The physical order gap rotates with the initial heading θ, so for a fixed goal the better order changes from scene to scene with the object's visible orientation. Both actions are linear in the exact state features, so the exact embedding remains a positive control. The textured F shows its heading through its shape; the photograph disc shows it only through the orientation of the photograph's content. Protocol, decoder and controls are those above, with 1,700 training and 300 test scenes in each world and layers chosen on validation by composed fidelity.

| World | Representation (layer) | Heading decodability $R^2$ | Fidelity: direct / composed / shared / specialised | Wrong choices: direct / composed | Relative regret: direct / composed |
|---|---|---|---|---|---|
| Both | Exact embedding | 1.00 | 1.00 / 1.00 / 1.00 / 1.00 | 0.0% / 0.0% | 0.000 / 0.000 |
| Both | Image-free reference (mean training gap) | – | – | 50.2–51.1% | 1.01–1.02 |
| Textures | SigLIP (L8) | 0.86 | 0.48 / 0.45 / 0.46 / 0.46 | 7.4% / 7.0% | 0.044 / 0.040 |
| Textures | DINOv2 (L12) | 0.34 | 0.13 / 0.08 / 0.08 / 0.05 | 27.9% / 30.3% | 0.454 / 0.498 |
| Textures | CLIP (L6) | 0.29 | 0.17 / 0.10 / 0.09 / 0.08 | 29.0% / 29.4% | 0.491 / 0.510 |
| Textures | MAE (L12) | 0.07 | 0.01 / 0.01 / 0.00 / −0.01 | 38.8% / 44.0% | 0.731 / 0.842 |
| Photo discs | DINOv2 (L8) | 0.21 | 0.16 / 0.14 / 0.15 / 0.13 | 32.8% / 33.1% | 0.594 / 0.590 |

| World | Representation (layer) | Heading decodability $R^2$ | Fidelity: direct / composed / shared / specialised | Wrong choices: direct / composed | Relative regret: direct / composed |
|---|---|---|---|---|---|
| Photo discs | MAE (L12) | 0.16 | 0.11 / 0.11 / 0.11 / 0.09 | 36.2% / 35.4% | 0.679 / 0.646 |
| Photo discs | SigLIP (L6) | 0.16 | 0.07 / 0.05 / 0.08 / 0.07 | 36.6% / 37.2% | 0.671 / 0.702 |
| Photo discs | CLIP (L6) | 0.14 | 0.12 / 0.11 / 0.11 / 0.09 | 37.1% / 36.6% | 0.683 / 0.675 |

The witness, which decodes the two true final states, errs on 6.2–8.9% of choices on textures and 2.1–2.8% on photographs at these layers.

- **The image now matters.** The image-free reference is at chance, while the witness chooses well. Neither the position decoder ($R^2$ 0.87–0.97) nor the representation of the outcomes limits the test.
- **Only SigLIP on textures chooses well from the initial state:** 7% wrong choices, close to its witness (6%). Every other encoder errs on 28–44% of choices with every predictor, better than chance but far from its witness.
- **Composition is not the bottleneck here.** At the validation layers the four predictors differ by 0.4 to 5.8 points, and the direct predictor, trained on the effect of both orders, helps little.
- **A control with known geometry accounts for the behaviour.** We decode the heading linearly from the initial representation, apply the simulator's law $\Delta p = d(R_{90} - I)\hat{u}$, and choose the order. This control uses the simulator's law, so it is diagnostic, not a competitor. At every layer of both worlds, the direct predictor's error lies within −1.0 to +5.5 points of this control, and within ±1 point on photographs. The composed predictor lies within −1.3 to +8.6 points of it (largest gap: DINOv2 layer 8 on textures). These aggregates show a strong association between choice error and heading readability; they do not establish a limit for every affine predictor, nor identical decisions scene by scene.
- **The heading is the limit.** Across all 48 encoder layers of both worlds (layers 2 to 12), the choice error of the direct predictor falls with the linear decodability of the heading ($\cos\theta$, $\sin\theta$) from the initial representation: Spearman $\rho = -0.98$, and −0.95 for the composed predictor (Figure 3). The relation holds without SigLIP (−0.98 and −0.94, 36 layers), on textures alone (−0.99 and −0.96) and on photographs alone (−0.98 and −0.86). Within each encoder and world, across its six layers, $\rho$ ranges from −0.89 to −1.00. Layers of one network are not independent, so these are descriptive robustness checks, not 48 replications. Because the physical target Δp is itself linear in the heading, the relation is expected for affine read-outs; the finding is where the deficit lies, not the correlation.

- **Two kinds of difficulty.** On textures the F's shape shows the heading. On photograph discs the heading must be inferred from the orientation of the photograph's content, which can be ambiguous even for a human observer. SigLIP makes the heading linearly readable on textures ($R^2$ 0.77–0.86 from layer 6); on photographs no encoder exceeds 0.23.

The encoders represent where the object is very accurately but, with one exception, not linearly which way it faces. When the better order depends on heading, the tested affine predictors do about as well as the heading information they can read allows. A planner built on these frozen features would need either a non-linear predictor or features that keep orientation.

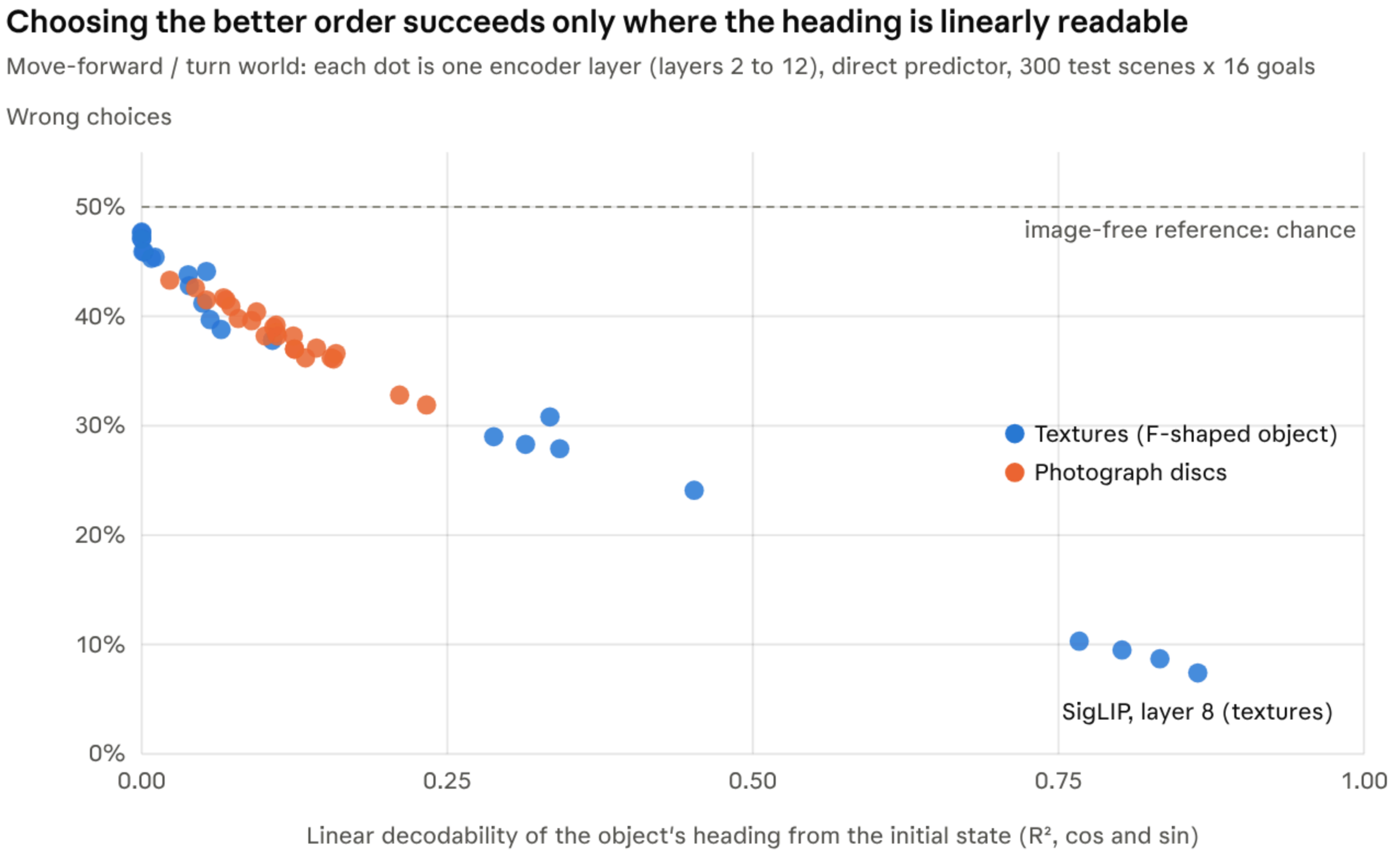

Figure 3 · results/fp_{texture,disc}_layers.json, 4 encoders × layers 2–12 × 2 worlds; negative decodability shown at 0

## A trained world model: two pushes in Push-T

The operators above are fitted separately on frozen features. To see whether the same gap between fidelity and decisions appears in a model trained end to end for planning, we took the two pretrained Push-T world models released with JEPA-WMs (Terver et al., 2025): JEPA-WM and the DINO-WM baseline, both on a frozen DINOv2 ViT-S/14 encoder with an action-conditioned predictor (facebook/jepa-wms on Hugging Face; code at commit 13cf1d9). We used their predictors unchanged, with the released normalisation statistics and one model step per five simulator steps, as in training. Each of 300 scenes places the pusher 55–95 px from the T block; move A points roughly at the block, move B turns 60–180° away, each lasting one model step. The simulator gives the four true observations after A, B, A then B, and B then A. In 209 scenes the two orders leave the block at least 2 px apart (median 24.6 px); the other 91 have no order effect and are reported separately.

Three conditions mirror the reviewer's protocol: the model rolls out both moves from the start frame (composed); it predicts only the second move from the encoded true intermediate frame (true intermediate); or the two true final frames are encoded (true finals, an empirical reference for the representation). A positive control comes first: one predicted move explains 68% (JEPA-WM, [66, 70]) and 75% (DINO-WM, [73, 76]) of the latent change that copying the start frame misses.

| Model | Gap fidelity, composed | Gap fidelity, true intermediate | Goal-image identification correct, scenes with an order effect (n = 209) | Same, block ends in the same pose (n = 82) | Position regret, composed | Position regret, true intermediate | Position regret, true finals | Regret, composed − true intermediate |
|---|---|---|---|---|---|---|---|---|
| JEPA-WM | 0.473 [0.428, 0.515] | 0.639 [0.609, 0.668] | 97.1% [95.5, 98.6] | 97.0% [93.9, 99.4] | 0.226 [0.186, 0.272] | 0.160 [0.123, 0.203] | 0.151 [0.113, 0.194] | +0.067 [+0.022, +0.111] |
| DINO-WM | 0.498 [0.443, 0.549] | 0.715 [0.691, 0.737] | 96.7% [95.0, 98.1] | 84.8% [79.9, 89.6] | 0.232 [0.183, 0.284] | 0.154 [0.117, 0.196] | 0.178 [0.136, 0.224] | +0.079 [+0.043, +0.116] |

Order identification uses the planners' own objective: the goal is the true final image of one order, and the model picks the order whose predicted final latent is nearer (visual L2 plus 0.1 × proprioceptive L2, as in their planner). Chance is 50%. The control column repeats the test on the 82 scenes where the block ends at the same position (within 0.5 px) and orientation (within 0.01 rad) in both orders, so that only the pusher and finer traces of the trajectory distinguish the two final images. The position test follows Section 6: a linear decoder of block position, fitted on encoded observed frames of held-out scenes ($R^2$ 0.98), and 16 goals around the midpoint of the two true outcomes; regret is the mean excess distance to the goal relative to chance (1 = chance). Intervals are 95% scene-level bootstrap intervals, paired for differences.

- **Fidelity is moderate, and observing the intermediate state recovers a substantial part of the deficit.** Composed rollouts recover about half of the latent order gap. Starting from the true intermediate frame raises fidelity by 0.17 (JEPA-WM, [0.12, 0.21]) and 0.22 (DINO-WM, [0.17, 0.26]), that is 31% and 43% of the deficit to a fidelity of 1.
- **High identification does not show that the block outcome is predicted.** Both planners identify which order produced a goal image in about 97% of scenes with an order effect. JEPA-WM does as well (97.0%) when the two orders leave the block in the same pose (within 0.5 px and 0.01 rad), and DINO-WM still reaches 84.8%. The final pusher positions of the two orders differ by about 5 px in both kinds of scene, and the planner score includes proprioception; with visual L2 alone, JEPA-WM falls to 74% on the identical-outcome scenes. Identification therefore rests largely on recognisable traces of the trajectory; how much of it rests on the block in the other 209 scenes remains to be measured.
- **The intermediate error has a measurable physical cost.** On block-position goals, composed rollouts leave a regret of 0.23 for both models. Giving the true intermediate frame lowers it by 0.067 (JEPA-WM) and 0.079 (DINO-WM), about 30% and 34%, close to the true finals (differences +0.009 [−0.016, +0.031] and −0.024 [−0.048, 0.000]). Composed regret exceeds the true-finals regret by 0.075 [0.026, 0.122] and 0.055 [0.013, 0.096]. Choice error rates move less (JEPA-WM 20.9% composed, 18.0% true intermediate, 17.4% true finals; DINO-WM 20.0%, 17.6%, 19.8%), because an error rate counts a near-tie inversion like a costly one.
- **The models know when order does not matter.** In the 91 scenes without an order effect, the decoded predicted gap has median 2.4–2.6 px, against 20 px in scenes with one.
- **Difficulty follows the size of the effect.** For JEPA-WM, composed choice errors fall from 31% for gaps under 19 px to 13% above 30 px (true finals: 28% to 9%).

**Goals on the block pose alone, on 300 fresh scenes.** To remove the pusher from the comparison, we drew 300 new scenes (seed 1; 221 with an order effect) and scored decisions on the block's pose only. A linear decoder of block position and orientation (x, y, 50 cos θ, 50 sin θ; $R^2$ 0.97 on held-out observed frames) reads each final latent. Targets are 16 block poses around the midpoint of the two true outcomes, in position–orientation space with 50 px per radian. Goals and physical costs depend only on block pose; the pusher remains visible to the encoder. Because the goals are placed around the two true outcomes, the protocol tests the choice between two sequences for controlled goals. Before any comparison, the score must rank the two true outcomes correctly: from the encoded true finals it does so in 80% of choices, an empirical reference for a linear block decoder fitted on these data, not a proven bound.

| Model | Regret, composed | Regret, true intermediate | Regret, true finals | Composed − true intermediate | Composed − true finals | Block-only identification error, composed | Block-only identification error, true finals |
|---|---|---|---|---|---|---|---|
| JEPA-WM | 0.315 [0.264, 0.370] | 0.234 [0.194, 0.278] | 0.192 [0.154, 0.236] | +0.081 [+0.042, +0.125] | +0.123 [+0.077, +0.171] | 18.1% [14.9, 21.3] | 13.3% [10.4, 16.5] |
| DINO-WM | 0.263 [0.211, 0.319] | 0.198 [0.159, 0.243] | 0.185 [0.145, 0.232] | +0.065 [+0.022, +0.114] | +0.078 [+0.033, +0.127] | 14.5% [11.3, 17.4] | 13.1% [10.2, 16.3] |

The cost of the intermediate error replicates on block-pose goals: observing the true intermediate state lowers regret by about a quarter (26% for JEPA-WM, 25% for DINO-WM). The remaining difference from the true finals differs between models: +0.013 [−0.010, +0.037] for DINO-WM, compatible with a small difference in either direction, and +0.042 [+0.012, +0.071] for JEPA-WM, a gap that remains. Identification changes once the pusher is removed: recognising the order from the decoded block pose alone, the composed predictions err in 15–18% of cases, against 3–5% with the planner score and 13% for the true finals. On these fresh scenes the planner score again identifies the order in 89% (JEPA-WM) and 78% (DINO-WM) of the 76 scenes where the block ends in the same pose (same tolerances), which confirms the earlier control. On these 76 scenes, block-pose identification accuracy was 50%, compared with 89% for JEPA-WM and 78% for DINO-WM using the planner score.

Push-T thus separates two abilities. The planners recognise which trajectory produced an image in 95–97% of cases, and in 78–97% even when the block outcome is identical, yet the error accumulated at the first predicted step still costs a quarter to a third of the regret on goals defined by the block, and observing the intermediate state removes most or all of that cost. Recognising a trajectory and predicting the consequence that matters for the goal are different, and a world model should be evaluated on the second. One design caution: our first version compared predictions with goal images rendered by us, whose block angle and pusher position differed from the true outcomes; every condition, the true finals included, was then at chance. Goals must be reachable states, or be read out through a task-relevant decoder.

# 7 Layer profiles: decodability is not compositional fidelity

The table gives, for each encoder, its validation-selected layer and its output layer. Composition and fidelity are for the composed operators.

| Setting | Encoder | Layer | Pose decodability $R^2$ | One-step $R^2$ | Composition | Fidelity r,tx | Fidelity sc,tx |
|---|---|---|---|---|---|---|---|
| Fully visible photos | DINOv2 | 6 / 12 | 0.99 / 0.98 | 0.92 / 0.74 | 0.076 / 0.217 | 0.81 / 0.41 | −0.43 / −0.76 |
| Fully visible photos | CLIP | 6 / 12 | 0.99 / 0.98 | 0.91 / 0.79 | 0.074 / 0.185 | 0.78 / 0.49 | −0.12 / −0.42 |
| Fully visible photos | SigLIP | 4 / 12 | 0.99 / 0.95 | 0.89 / 0.55 | 0.109 / 0.419 | 0.72 / 0.22 | 0.13 / −0.21 |
| Fully visible photos | MAE | 8 / 12 | 0.99 / 0.99 | 0.87 / 0.85 | 0.104 / 0.131 | 0.66 / 0.64 | −0.03 / −0.08 |
| Fully visible photos | Pixels 24×24 / RGB PCA-128 | – | – | 0.96 / 0.96 | 0.096 / 0.074 | 0.83 / 0.88 | 0.46 / 0.66 |
| Textures | DINOv2 | 6 / 12 | 0.89 / 0.91 | 0.71 / 0.54 | 0.271 / 0.451 | 0.57 / 0.19 | 0.21 / 0.03 |
| Textures | CLIP | 4 / 12 | 0.87 / 0.87 | 0.64 / 0.06 | 0.332 / 1.014 | 0.46 / 0.04 | 0.15 / −0.03 |
| Textures | SigLIP | 4 / 12 | 0.92 / 0.93 | 0.74 / 0.49 | 0.240 / 0.487 | 0.59 / 0.21 | 0.25 / 0.03 |
| Textures | MAE | 2 / 12 | 0.76 / 0.86 | 0.29 / 0.41 | 0.735 / 0.566 | 0.10 / 0.12 | −0.04 / −0.06 |
| Textures | Pixels 24×24 / RGB PCA-128 | – | – | 0.29 / 0.25 | 0.761 / 0.823 | 0.10 / 0.13 | −0.07 / 0.05 |
| Window photos | DINOv2 | 2 / 12 | not identifiable | 0.69 / 0.24 | 0.336 / 0.726 | 0.28 / 0.03 | 0.10 / −0.02 |
| Window photos | CLIP | 0 / 12 | not identifiable | 0.79 / 0.57 | 0.237 / 0.472 | 0.45 / 0.10 | 0.22 / 0.02 |
| Window photos | SigLIP | 0 / 12 | not identifiable | 0.80 / 0.40 | 0.219 / 0.568 | 0.42 / 0.01 | 0.23 / −0.02 |

| Setting | Encoder | Layer | Pose decodability $R^2$ | One-step $R^2$ | Composition | Fidelity r,tx | Fidelity sc,tx |
|---|---|---|---|---|---|---|---|
| Window photos | MAE | 0 / 12 | not identifiable | 0.85 / 0.62 | 0.177 / 0.351 | 0.53 / 0.27 | 0.32 / 0.10 |
| Window photos | Pixels 24×24 / RGB PCA-128 | – | – | 0.90 / 0.88 | 0.144 / 0.162 | 0.63 / 0.61 | 0.60 / 0.39 |

Reading.

- **The pose stays decodable where composition degrades.** On fully visible photographs, pose decodability is 0.95–0.99 at the output of all four encoders, while composed fidelity for scale and translation is negative in all four.
- **Where the pose is visible, the most compositional layers are in the middle of the network.** DINOv2 and CLIP peak at layer 6 and SigLIP at layer 4; MAE is flat from layer 2 to 12. On textures the peak is at layers 4 to 6, and CLIP's operators stop composing after layer 6 (composition defect above 1).
- **In the window design, fidelity is highest at the input and decays with depth in all four encoders.** An earlier version reported a monotonic decay and a reversed ranking of encoders on photographs. We now attribute both to partial observation.
- **Pixels are not a ceiling.** On textures, where the object is textured and the background cluttered, every encoder except MAE beats pixels on every defect at its best layer. On photographs, pixels compose better.

In the grey shape world at 112×112, DINOv2's patch embedding (layer 0) is slightly more compositional than pixels (fidelity r,tx 0.71 against 0.65) while its decodability is only 0.60. From layer 2, decodability reaches 0.91 and fidelity drops to 0.29, then to about 0 at layers 10–12. At 224×224 the peak moves to layers 6–8 (0.47–0.48, against 0.77 for pixels). Global tokens (class token, mean patch token) do not encode the pose (decodability −0.05 to 0.05), so their defects reflect invariance and are not interpreted.

# 8 Language models

Models and worlds. We use SmolLM3-3B-Base (Hugging Face, 2025; 36 layers) and Qwen3-0.6B-Base (Qwen Team, 2025; 28 layers), run on Apple MPS.

- **Days world:** "[context]. The appointment is on Tuesday." Transformations are +1, +2, +3 days and their inverses (group $Z_7$, commutative).
- **Shelf world:** three nouns on a shelf. Transformations are two swaps and two rotations (group $S_3$, non-commutative).

Contexts are random short sentences, so a transformation changes only the group element. We read the residual stream at every third layer, with PCA 64, on 600 training and 200 test prompts per transformation set.

Behaviour. SmolLM3 solves one-hop day arithmetic (“If today is Monday, then two days later it will be”) at 96% and two hops at 36% (chance 14%): a clear two-hop curse. Qwen3-0.6B is near chance (31% and 29%). Behavioural tests therefore use SmolLM3, with Qwen3 as a replication where its baseline allows.

A day that is mentioned. At the last token of the days world, the day stays linearly decodable deep into SmolLM3 while its shift operators stop composing:

| SmolLM3 layer | Day decodability $R^2$ | One-step $R^2$ | Inverse (median) | Composition | Predicted commutator (commuting pairs) |
|---|---|---|---|---|---|
| 3 | 1.00 | 0.98 | 0.01 | 0.02 | 0.001 |
| 6 | 0.98 | 0.90 | 0.04 | 0.11 | 0.007 |
| 12 | 0.97 | 0.86 | 0.07 | 0.14 | 0.007 |
| 18 | 0.92 | 0.65 | 0.19 | 0.32 | 0.020 |
| 24 | 0.77 | 0.42 | 0.30 | 0.55 | 0.019 |
| 30 | 0.84 | 0.37 | 0.34 | 0.59 | 0.012 |
| 36 (output) | 0.57 | 0.12 | 0.41 | 0.90 | 0.011 |

Qwen3-0.6B shows the same early algebra (layer 3: one-step $R^2$ 0.98, inverse 0.01, composition 0.02), with a later and milder loss (composition 0.18 at layer 15, 0.45 at 24, 0.41 at the output). In both models, predicted commutators stay at most 0.03 of the move energy at every layer, as the commutative world requires.

A day that is computed. At the last token of a one-hop prompt, where the model writes its answer, we fit operators A_b that map the state for day D to that for D+b. We then edit the state as h + U(A_b z − z).

- **From layer 28 in SmolLM3, the edit shifts the answer by b at 64–73%.** This is within a few points of injecting the true state (oracle) and of the unpatched accuracy (73.8%), against 9–21% for operators fitted on shuffled pairs.
- **The operators compose.** Applying $A_1$ then $A_2$ gives the +3 answer at 63–67%, matching the directly fitted $A_3$.
- **Before layer 22, neither the operators nor the oracle move the answer,** so early layers say nothing about the operators' causal role.

- **Qwen3 replicates at layers 22–25** (operators 0.37–0.42, oracle 0.38–0.45, composed 0.37, shuffled control 0.16–0.20), within the limit of its weak baseline.

This geometry is approximately, not exactly, cyclic.

- Applying $A_1$ seven times and injecting the result leaves the prediction unchanged on 83–90% of items.
- Accuracy changes by −0.02 [−0.06, +0.02] at layer 25, and by −0.10 to −0.13 at layers 28–34.
- The latent residual $\|A_1^7 z - z\|^2$ is 0.12–0.23 of the energy of a one-day move.

A centroid edit, which moves the state between the mean states of the current and target answers, does nearly as well as the fitted operators on the training template. The intervention therefore shows an editable geometry of the seven answers that respects the cycle. It does not show that the operators carry more structure than that geometry.

Two-hop prompts. Applied to two-hop states, the one-hop operators carry shift-specific information: from layer 28 they beat a wrong-shift control by +0.08 to +0.13. They move the state far less effectively than a state that carries the answer, and without a context sentence they lower overall accuracy, because they break answers that were already right. They are a partially causal linear summary, not the model's own mechanism. Details, the shelf world and back-patching are in Appendices A and D.

# 9 Transfer between encoders

If two encoders were related by an invertible affine map M on the data, their operators would be conjugate, $A_2 = M A_1 M^{-1}$, and the two paths $M A_1 z$ and $A_2 M z$ would coincide. We test this directly, without a pseudo-inverse, for every ordered pair of encoders at their validation-selected layers. M is fitted on all training images, or with the tested action's images held out; no score changes by more than 0.01. We score each path's predicted move, path − M z_a(x), against the true move in the target encoder, z_b(gx) − z_b(x):

```
R^2_{\mathrm{disp}} = 1 - \frac{\sum \lVert (\mathrm{path} - M z_a(x)) -
(z_b(gx) - z_b(x)) \rVert^2}{\sum \lVert z_b(gx) - z_b(x) \rVert^2}, \qquad
\mathrm{agreement} = 1 - \frac{\sum \lVert M A_1 z - A_2 M z \rVert^2}{\sum
\lVert z_b(gx) - z_b(x) \rVert^2}
```

Agreement measures how consistent the two paths are with each other; displacement $R^2$ measures how accurate each is. The map-only line, M z_a(gx) − M z_a(x), is a reference, not a ceiling.

| Setting (12 ordered pairs) | Agreement of the two paths | Displacement $R^2$: transform then map | Displacement $R^2$: map then transform | Displacement $R^2$: map only | Displacement $R^2$: native operator | Static error of M (share of move energy) |
|---|---|---|---|---|---|---|
| Fully visible photographs | 0.95–0.98 | 0.76–0.86 | 0.79–0.87 | 0.75–0.86 | 0.87–0.92 | 0.11–0.27 |
| Textures | 0.72–0.94 | 0.22–0.66 | 0.30–0.71 | 0.17–0.82 | 0.29–0.74 | 0.85–11.5 |

- **Fully visible photographs.** The two paths agree at 0.95–0.98, and both transfer the move nearly as well as the target's native operator. Scored on endpoints instead (0.70–0.87), they look worse only because of the static error of the point map, 11–27% of the move energy.
- **Textures (350 training scenes).** The point map is poor: its static error exceeds the move, and endpoint scores are strongly negative. The moves still transfer partially, and the two paths still agree at 0.72–0.94.

An RBF kernel map, tested under the earlier per-axis normalisation, did not align the encoders better than a linear one.

Relation to the Platonic Representation Theorem. Balestriero and LeCun (2026) prove that JEPAs trained on several modalities with good views learn the same representation up to a rotation, in an idealised setting. They state that action-conditioned predictors are outside their analysis. If two encoders were equal up to a rotation Q, every operator would transfer exactly. Our encoders meet none of the theorem's hypotheses, so our results are compatible with it. They add an operator-level reading: where the pose is visible, encoders agree on how a transformation moves a state far more closely than global point-alignment metrics suggest. The open question becomes whether encoders whose points align at the scale of a single move also share their operators.

# 10 Theory: the commutator is the other half of the counterfactual coupling

Let the state follow $\dot{s} = v_a(s) + b_\varepsilon(s)$, with $v_a$ the effect of the action and $b_\varepsilon$ that of a hidden disturbance, both held fixed over a step h (Balestriero et al., 2026). Changing the action from a to a′ changes the field by $\delta v = v_{a'} - v_a$. Write $K_\varepsilon = D\delta v \cdot b_\varepsilon + Db_\varepsilon \cdot \delta v$. To second order, the disturbance-dependent part of the action contrast is

```
\Delta_h(\varepsilon) - \mathbb{E}\,\Delta_h = \tfrac{h^2}
{2}\big(K_\varepsilon - \mathbb{E}\,K_\varepsilon\big) + O(h^3)
```

Its variance over $\varepsilon$ is the $O(h^4)$ residual-reuse error of the counterfactual paper. When $b_\varepsilon$ is linear in a zero-mean Gaussian $\varepsilon$, $E\,K_\varepsilon = 0$ and the expression reduces to $(h^2/2)\,K_\varepsilon$. Applying the two influences one after the other instead gives the order gap

```
\varphi^{h}_{b}\circ\varphi^{h}_{\delta v}(s) - \varphi^{h}_{\delta
v}\circ\varphi^{h}_{b}(s) = h^2\big(D b\, \delta v - D\delta v\, b\big) +
O(h^3)
```

which is the Lie bracket at leading order. Both quantities are built from the same two cross-derivative terms, $D\delta v \cdot b$ and $Db \cdot \delta v$: the counterfactual interaction uses their sum, the commutator their difference. Sum and difference reconstruct the two signed vector terms only when both are observed at the same state, which a scalar variance and a scalar fidelity are not. The Lie bracket is the infinitesimal leading term; the empirical gaps of Sections 4 to 9, at 4 to 50 px, are finite-step quantities.

The two halves dissociate. We integrated $ds/dt = a \cdot u(s) + \varepsilon \cdot b(s)$ with RK4 for four pairs of fields, fitting slopes in log-log over h from 0.05 to 0.4:

| Pair u, b | Symmetric half (counterfactual) | Antisymmetric half (commutator) | Residual-reuse ratio at h = 0.05 (slope) | Order gap at h = 0.05 (slope) |
|---|---|---|---|---|
| Translation, translation | 0 | 0 | 3e−27 (machine zero) | 7e−29 (machine zero) |
| diag(1,−1)s, [[0,1],[1,0]]s (anticommuting) | 0 | ≠ 0 | 5.3e−7 (3.9) | 4.9e−3 (1.9) |
| Scaling s, rotation Js | ≠ 0 | 0 | 2.4e−3 (1.8) | 5e−29 (machine zero) |
| Rotation Js, translation | ≠ 0 | ≠ 0 | 6.3e−4 (2.0) | 8.8e−4 (2.0) |

Scaling and rotation commute exactly, yet residual reuse fails at second order. Anticommuting linear fields do not commute, yet residual reuse is exact to second order. Neither test certifies the other.

Proposition. Suppose that on the data support the encoder represents the transformations affinely, $f(T_g x) = \rho(g) f(x) + c_g$, with homogeneous matrices satisfying $\tilde{\rho}(hg) = \tilde{\rho}(h) \tilde{\rho}(g)$, and that the augmented features $(z, 1)$ span their space. Then unregularised least squares recovers $\tilde{A}_g = \tilde{\rho}(g)$, the inverse and composition defects are 0, and the predicted order gap equals the observed one: zero where the world commutes, exact where it does not.

The converse does not hold: zero defects do not by themselves make $g \mapsto A_g$ a homomorphism. Consider the linear case with zero biases and take $A_g = -\rho(g)$. Then $A_h A_g = \rho(hg)$, so inverse, two-step and order-gap predictions are exact, whereas $A_{hg} = -\rho(hg) \neq A_h A_g$ and $A_e = -I$. A converse needs in addition exact one-step prediction by every $A_k$, including $k = hg$, on a support that contains the relevant intermediate states; then $A_h A_g z = A_{hg} z$ on that support. Faithfulness further requires injectivity: transformations that act identically on the support cannot be told apart.

# 11 Discussion and limitations

What the results establish. Two claims, of different strength. First, in all 24 paired contrasts, composing actions learned on initial states predicts the order effect worse than a direct read-out from the initial representation. Second, in 23 of the 24, that read-out also beats an image-free constant, so the order effect is predictable from scene-specific information in the initial representation; MAE on textures, for rotation and translation, is the exception. The composition deficit has two parts:

- a coverage part, recovered by training each action on a mixture of initial and transformed states;
- a conditioning part, recovered by fitting the second action on states produced by the first.

Their shares depend on the transformation pair and on what the observation shows. In the two-action choice test where the better order depends only on the goal, both fidelity and regret reward the coverage fix, but only fidelity clearly distinguishes the step-specific fix. Where the better order depends on the object's heading, composition stops being the bottleneck: all four predictors are limited alike by how linearly the encoder exposes the heading.

What it means for world models. Planning with learned actions assumes that they compose. Our results give a cheap check, the 24-contrast protocol with its image-free reference, which can be run on any frozen encoder and any action-conditioned predictor. In the setting we tested they also give a cheap remedy: training each action on real intermediate states as well as initial ones. Rollout training exposes a predictor to its own predicted intermediate states, which follow a different distribution; whether it brings the same benefit is a direct next test. Finally, a fidelity gain need not change a decision: step-specific fitting gained 0.13–0.16 in fidelity with little change in regret for the tested goals.

Open questions.

- **State-dependent choices beyond affine read-outs.** In the move-forward and turn world, the witness on the true final states chooses well while affine predictors from the initial state do not, except where the heading is linearly readable. Non-linear action predictors on the same frozen features, or encoders trained to keep orientation, are the direct next tests.
- **Active learning.** Can an agent choose which transformed states to learn from, so as to compose better? The baseline to beat is no longer random sampling but the balanced mixture, which already takes DINOv2 from −0.429 to 0.649 on scale and translation. A strategy guided by uncertainty about order would have to beat that mixture at an equal interaction budget, on unseen plans.

Limitations.

- Affine operators in a PCA space of 128 (vision) or 64 (language) dimensions are the simplest hypothesis; a representation may carry the algebra in a curved chart.
- There is one base-size model per training objective, and two language models.
- Two non-commuting pairs were tested, with actions about the image centre.
- Defects are finite-step quantities and mix the encoder's geometry with the step size.
- Fidelity is undefined when the true gap is zero and unstable when it is small. Steps were chosen so that the gap is at least about 10% of the move energy.

# Appendix A: back-patching, all series

We copied the last-token state of a two-hop prompt from a later layer s into an earlier layer t of a second pass on the same prompt (Biran et al., 2024). The control injects the state of another prompt with a different answer. Pairs were sampled every three layers (78 pairs) on the first item set, and the two best were re-tested.

| Run and item set | 16 → 13 | 19 → 10 |
| --- | --- | --- |
| First run, seed 0, 210 items (selection): patched / control | 36.7% / 29.0% | 33.3% / 31.9% |
| Second run, seed 1, 420 items: patched / control | 38.8% / 36.0% | 40.0% / 31.2% |
| Third run, the original 210 items: patched − control, 95% CI | +0.076 [+0.029, +0.124] | +0.014 [−0.024, +0.052] |
| Third run, 420 fresh items: patched − control, 95% CI | +0.086 [+0.050, +0.121] | −0.010 [−0.045, +0.024] |

Each run finds a pair above its control, but not the same pair every time. We report this instability as a result: back-patching effects of this size are not a reliable signal of where a second hop is computed.

# Appendix B: normalisation

Per-axis standardisation of PCA coordinates, used in an earlier version, gives low-variance coordinates the same weight as principal ones.

- On fully visible photographs it lowers composed fidelity on rotation and translation from 0.79 to 0.38 for DINOv2 at layer 8, and from 0.75 to 0.29 for CLIP at layer 4.
- It breaks the pixel positive control (one-step $R^2$ −43 instead of 0.98), because constant background pixels survive PCA as near-zero-variance coordinates.
- In language models, the global scale lowers the early inverse defect (SmolLM3, layers 3–6: 0.01–0.04 against 0.09–0.19) and leaves the depth trend and the shelf result unchanged.

Every direction we checked survives the change; levels do not.

# Appendix C: compute and reproducibility

Vision features were extracted on Apple MPS (about 10 minutes per encoder for 2,000 photographs) and analysed on CPU. PCA was fitted on 600 training images. The archive that accompanies this preprint distributes the scripts (42 Python files) and every aggregated result (101 JSON files, including per-item predictions for the language-model tests); it does not include the saved PCA activations (about 1.2 GB). These are regenerated deterministically from the public model weights and Imagenette with feats_all.py (texture, window and disc worlds) and feats_fp.py (move-forward and turn world); the launch commands, seeds, splits, step sizes and PCA settings are given in the scripts' headers. The analyses are review2_analysis.py (defects, order tests, ridge sweep, affine control), baseline_mean.py (image-free reference and hybrid predictor), nonlin_shared.py (shared non-linear operators), decision_test.py, fp_analysis.py, fp_layers.py and fp_heading_control.py, intertwine2.py, run_llm_review2.py and llm_review2_analysis.py. The Push-T pilot uses pusht_order_v3.py, pusht_analysis.py and pusht_blockpose.py (block-pose goals; fresh scenes launched by lancer_pusht6.command) (earlier versions pusht_order.py and pusht_order_v2.py are kept for the record), small stand-ins for the gym, decord and lpips imports that the simulator and model code do not use at inference (stubs.tgz), and the launcher lancer_pusht4.command; the world-model weights are downloaded from facebook/jepa-wms.

# Appendix D: language models, further results

In the shelf world the arrangement is poorly decodable from the final token ($R^2$ 0.42 at layer 3 in SmolLM3, below 0.15 from layer 21). Commutator fidelity is at most 0.16 at every layer in both models. On the nouns' own tokens the $S_3$ algebra is near exact (fidelity 0.84–1.00), but that is built in: permuting nouns whose token states depend mainly on the noun permutes blocks of the concatenated vector.

Two-hop prompts. We applied the one-hop operators to the last-token state of 420 two-hop prompts, with and without a context sentence, using A_b for the actual second shift, and compared with the same edit for a wrong shift. A positive control comes first: injecting the state of a one-hop prompt whose correct answer is the two-hop target gives 0.70 accuracy with context and 0.94–0.96 without, from layer 28 (unpatched 0.09 and 0.32). The pipeline can therefore write an answer when it is given a state that carries one.

| Without context | L28 | L31 | L34 |
|---|---|---|---|
| Unpatched accuracy | 0.32 | 0.32 | 0.32 |
| Operator A_b | 0.13 | 0.21 | 0.23 |
| Operator − wrong-shift control, 95% CI | +0.09 [+0.05, +0.12] | +0.13 [+0.08, +0.17] | +0.11 [+0.06, +0.15] |
| Centroid edit | 0.23 | 0.24 | 0.25 |

| Without context | L28 | L31 | L34 |
|---|---|---|---|
| First-hop answers converted, operator (centroid) | 13% (51%) | 55% (58%) | 42% (58%) |
| Answer-carrying state (positive control) | 0.96 | 0.94 | 0.96 |

With context, at layers 28–34, the operator edit adds +0.07 to +0.09 over unpatched and beats the wrong-shift control by +0.08 to +0.10 (intervals exclude zero); the centroid edit adds +0.20 to +0.22.

- The operators carry shift-specific information into two-hop states, but they move the state far less effectively than a state that carries the answer.
- Without context they lower overall accuracy, because they break answers that were already right (94–119 right-to-wrong transitions, against 23–73 wrong-to-right).

They are a partially causal linear summary, not the model's own mechanism. Back-patching later layers into earlier ones gave a positive pair in each run, but not the same pair every time (Appendix A).